# Humor Style Drives Laughter, Topic Shapes Acceptability: Evaluating Bilingual Personal and Political Robot-Delivered AI Jokes

Anna-Maria Velentza and Anne-Gwenn Bosser

*Univ Brest-Bretagne INP, COMMEDIA team, Lab-STICC CNRS UMR 6285, France*

***Abstract*—Humor plays a central role in human social relationships, and recent advances in computational humor create new opportunities for integrating humor into human-robot interaction (HRI). While large language models (LLMs) can generate diverse forms of humor, it remains unclear how humor style, joke content, and language preference shape perceptions of robot-delivered humor in group settings. In this exploratory study, we employed a mixed factorial design in which participants evaluated AI-generated jokes delivered by a robot in a university classroom. We examined the effects of humor type (Affiliative, Self-Enhancing, Aggressive, Self-Defeating) and joke content (person-related vs. political) on perceived funniness and appropriateness, as well as preferred language. Results show that humor type significantly influences funniness, with Aggressive and Affiliative humor rated higher, while joke content primarily affects appropriateness, with person-related jokes preferred over political ones. Language preference was shaped by both joke content and participants' self-reported fluency and humor practices.**

# I. Introduction

Computational humor is an established research topic in AI and computational linguistics, utilizing Natural Language Processing (NLP) and Natural Language Understanding (NLU) to develop algorithms for joke generation [1]. Recent large language models (LLMs) demonstrate a high capacity for generating various forms of humor. Humor is crucial in human relationships [2], often deployed in group settings to resolve conflicts [3]. Within Human-Robot Interaction (HRI), humor has been explored as a mechanism to increase engagement and perceived robot likability [4], [5]. However, humor is highly context-dependent. Jokes are evaluated differently when the person delivering them is considered a member of the listener's in-group [6], whereas humor is central to human efforts to cope with political struggles and is compelling in understanding power dynamics in socio-political life [7]. Humor can also serve functional roles in HRI, such as easing social tension, mediating interactions within groups, and supporting more natural long-term collaboration between humans and robots [8], [9].

Motivated by our focus on incorporating humor into HRI to support robots as engaging and relatable partners in group settings and to enable them to function as social intermediaries, interaction mediators, or active group members, we designed an exploratory study in which the humanoid robot Navel delivered AI-generated jokes to a group of students. We evaluated each joke in terms of perceived funniness and appropriateness. The jokes varied in thematic content, focusing on either personal or political topics. Given the importance of culture and language in humorous communication, we also examined language preference by comparing French and English, which served as the language of instruction in the institution they were studying and the medium of collaboration in an increasingly international academic environment.

Our most important findings suggest that (a) humor style significantly affects perceived funniness, (b) joke content primarily shapes perceived appropriateness, and (c) language preference depends on both content and participants' linguistic background.

# II. Related Work

## *A. Humor and AI*

Computational humor has long explored rule-based and linguistic approaches to joke generation before the emergence of neural language models [10]. Recent advances in LLMs have challenged the long-standing assumption that humor is uniquely human. Empirical studies show that AI-generated jokes can be rated comparably to human-authored ones, particularly under conditions of source blindness [11]. Models such as GPT-4o have demonstrated strong performance in sentence-level humor and in humor-based coping strategies, in some cases outperforming humans in responding to negative social situations [12]. These findings suggest that humor generation emerges from broader capabilities such as contextual reasoning and creative language use rather than humor-specific mechanisms alone [13], [14].

Despite these advances, limitations remain. AI systems often struggle with forms of humor that rely on linguistic ambiguity and cultural nuance, such as puns. In translation tasks, AI-generated humorous content is frequently perceived as less natural than human translations, tending toward literalism at the expense of preserving humorous intent [15]. Research in computational humor continues to evolve, supported by dedicated venues such as the JOKER shared-task series at CLEF, which supports the evaluation of the capacity of AI systems to locate, classify, generate, and translate humorous sentences [16].

Humor also plays a significant role in shaping human-machine interaction. Prior work indicates that humorous behavior in conversational agents increases user engagement, enjoyment, and perceived social connection [4], [17]. Humor has been explored in diverse application domains, including virtual assistants and mental health support systems, where it can enhance conversational quality and user receptivity [14], [18]. Neurophysiological studies further suggest that humans process AI-generated humor differently from human humor, even when subjective funniness ratings are similar [19].

### *B. Humor and Robots*

Within HRI, humor has been widely investigated as a mechanism for enhancing social presence, likeability, and engagement. A systematic review by Oliveira et al. [4] reports generally positive effects of humor on users' perception of robots and interaction quality, while also noting methodological limitations and inconsistencies across studies. Experimental work has shown that humor can improve perceived task enjoyment, robot personality, and speaking style, whereas empathy primarily influences evaluations of interaction smoothness and receptivity [20]. Using the humanoid robot Pepper, Zhang et al. found that self-defeating humor positively influenced joke ratings and was associated with increased happiness, as measured through both subjective assessments and physiological indicators [21]. Other work demonstrates that humorous robots can increase social attraction and reduce negative affect, including diminished disgust responses to disparaging jokes, underscoring both the power and the ethical sensitivity of humor in HRI [22].

Adaptivity has emerged as a key theme in robotic humor. Reinforcement learning approaches have been used to dynamically tailor a robot's humor based on real-time user reactions, such as laughter [23]. Personality differences also modulate humor perception: for example, preferences for self-ironic versus schadenfreude-based robotic humor vary with users' traits such as neuroticism and openness [5]. Recent systems integrate contextual sensing, user profiling, and emotional state modeling to generate situationally appropriate humor using LLM-based pipelines [24]. Importantly, humor is not universally beneficial. In scenarios involving robot failure, humor can mitigate negative perceptions in low-severity situations but may exacerbate frustration or distrust when failures are severe [25]. Applications of robotic humor span education [26], performative contexts such as robot comedy shows [27], [28], highlighting both its potential and its contextual dependence.

### *C. Present Study*

Despite progress in computational humor generation, little work has examined how AI-generated humor is perceived when delivered by an embodied robot within a real social context. Most evaluations focus on text-based humor generation rather than socially situated interactions. However, humor perception depends strongly on group dynamics, topic sensitivity, and cultural context [3], while social identity theory suggests that humor directed toward in-group members may reinforce group cohesion when interpreted as playful teasing [29]. Additionally, two dimensions remain understudied in HRI humor research: the social target of humor (e.g., jokes about group members versus societal topics) and language preference in multilingual environments.

To address these gaps, we conducted an exploratory study in which a humanoid robot delivered AI-generated jokes to a group of bilingual students in a shared social environment. The jokes varied along two key dimensions: humor style, following the four-style model of humor [30], and joke content, referring either to personal group-related topics or contemporary political situations. Participants evaluated each joke in terms of perceived funniness, appropriateness, and preferred language. By examining these dimensions simultaneously, the study aims to better understand how humor generated by AI is interpreted when embedded in a social robotic interaction.

The humor types used in the study follow the four-style model of humor proposed by Martin et al. [30], which distinguishes affiliative, self-enhancing, aggressive, and self-defeating humor (see Table I for description). This framework has been widely used in psychological research to study the social and interpersonal functions of humor.

### *D. Research Questions and Hypotheses*

**RQ1**: How does humor style influence the perceived funniness and appropriateness of robot-delivered jokes? **RQ2**: How does joke content influence the perceived funniness and appropriateness of robot-delivered jokes? **RQ3**: How do humor style, joke content, and language background influence preferred language for jokes?

**H1: Humor type will significantly affect perceived funniness.** Affiliative humor, which promotes social bonding, is expected to be rated as the funniest style, followed by aggressive humor, which may produce laughter through playful teasing within established social groups [3], [30].

**H2: Humor type will influence perceived appropriateness.** Affiliative and self-enhancing humor are expected to be perceived as more appropriate than aggressive or self-defeating humor, as the latter may involve social risks or negative self-presentation [30].

**H3: Joke content will affect perceived appropriateness.** Person-related jokes referring to familiar group members are expected to be perceived as more appropriate than political jokes, as political humor can involve sensitive social or ideological topics [7].

**H4: Joke content will have a weaker effect on funniness than humor type.** While topic may influence appropriateness, the humorous style itself is expected to be the primary determinant of perceived funniness.

**H5: Language preference will depend on both content and participant background.** Participants are expected to prefer jokes in the language that best aligns with their social context or cultural familiarity, particularly for jokes referencing shared social experiences or political events.

## III. Method

### A. Participants

Sixteen participants, aged 23-25 years old, were recruited from the university student population of the English-speaking engineering MSc program at the INP-Bretagne located in France (francophone) where the students were enrolled. They had normal or corrected to normal vision and hearing. The study is exploratory and aims to identify patterns that can inform larger-scale investigations. The study had been approved by the corresponding ethics committee.

### B. Design

The Navel robot [31] was introduced to the participants by the experimenter, who informed them that the robot would tell them some 'robot-generated' jokes. They became familiar with it by talking to it and receiving AI-generated jokes related to robots and technology use, which were generated with GPT-4. After familiarization and a detailed explanation of humor types, students were asked to evaluate a new set of jokes performed by the robot. The jokes had been curated and chosen among AI-generated jokes after giving as prompts keywords and the direction of the joke. All jokes were generated through prompt engineering specifying the desired humor style and topic. The prompts included instructions regarding tone, length, and thematic constraints. Human researchers reviewed and curated the generated outputs to ensure clarity, relevance, and alignment with the experimental conditions. This human-in-the-loop curation process was used to mitigate potential issues associated with uncontrolled AI generation. Paper-based responses were used to minimize interaction interference and maintain focus on the robot's performance. The robot delivered jokes using a neutral speaking style without adaptive timing or expressive gestures.

The study employed a mixed factorial design combining between-subject and within-subject factors. Between-subject factors were participants' self-reported *Fluency* (English, French, or Both) and *Preferred Joking Language* (English, French, or Both). Within-subject factors included *Humor Type* (Affiliative, Self-Enhancing, Aggressive, Self-Defeating) and *Joke Category* (Person vs. Politics). Each participant evaluated jokes from every combination of Humor Type and Joke Category, resulting in a fully crossed 4 x 2 within-subjects design.

### C. Joke Generation and Validation

To ensure adequate stimulus quality and coverage of the experimental conditions, a larger pool of candidate jokes was initially generated using GPT-4 through structured prompts specifying the target humor type (Affiliative, Self-Enhancing, Aggressive, Self-Defeating) and topic category (Person vs. Politics). Approximately four times the number of jokes required for the experiment were generated. From this initial pool, a subset of jokes was selected following an internal validation procedure.

Three independent researchers evaluated the candidate jokes to assess their correspondence with the intended humor type and topic category. Inter-rater agreement was assessed using Fleiss' k to ensure consistent classification. The resulting agreement indicated substantial reliability across raters (k = 0.84), confirming that the selected jokes clearly represented their intended humor categories.

### D. Contextualization of Humor Content

Prior to stimulus generation, an informal pilot workshop was conducted with the participants' cohort to identify shared humorous references within the student group. During this session, students discussed memorable or humorous moments involving their classmates, as well as personality traits or situations that were commonly joked about within the group. Importantly, individuals referenced in these anecdotes explicitly confirmed that the situations were perceived as humorous and acceptable within the social context.

The collected themes were then used as inspiration for generating person-related jokes with GPT-4. The generated jokes were not direct reproductions of students' own jokes but rather new variations based on the shared themes identified during the workshop. This procedure was intended to increase ecological validity by grounding the jokes in the participants' shared social context. The personal jokes per category in both languages can be found in Table I.

Political jokes were designed to reference the contemporary political situation in France to maintain relevance for participants. Humor research suggests that topical humor is often perceived as more engaging and relatable when it reflects shared cultural or political contexts [32]. The political jokes were carefully written to maintain a satirical tone without targeting specific individuals in a derogatory manner. Political satire often operates through aggressive humor mechanisms while simultaneously reinforcing shared attitudes within the audience, and thus jokes were tested to fit in the four categories correspondingly, as shown in Table II.

### E. Procedure

Participants evaluated eight jokes, each one classified by humor type (affiliative, self-enhancing, aggressive, self-defeating) and topic category (person, politics) and provided three responses for each joke: (1) whether the joke was *funny* (Yes/No), (2) whether the joke was *appropriate* (Yes/No), and (3) their *preferred language* for the joke (English, French, or Both). In total, the design allowed assessment of the effects of humor type and joke category within participants, while accounting for individual differences in language fluency and joking preference as between-subject factors. Forced-choice formats were chosen based on [33]'s meta-analysis showing that they are harder to fake than Likert scales. The robot was standing in the middle of the room, placed on a table, and in a randomized order, delivering the jokes, some first in French and some first in English. After each joke, the students were evaluating them on the given paper sheet.

### F. Group Interaction Context

Although participants evaluated jokes individually on paper, the study was conducted in a shared classroom environment in which the robot addressed the group collectively. This design choice was motivated by prior research indicating that humor is inherently social and often functions to reinforce group cohesion and shared identity [34], [35]. By presenting jokes in a group setting, the study aimed to approximate a naturalistic humor context and explore whether the robot could be perceived as a socially integrated participant within the group.

Group-based exposure to humor is known to influence laughter, evaluation, and perceived appropriateness due to social contagion and shared interpretive frameworks [35]. Therefore, maintaining a collective setting allowed the robot's humor to be experienced in a socially meaningful context rather than as isolated individual interactions.

## IV. Results

The within-subject design and repeated evaluations across multiple jokes increase the effective number of observations, allowing the detection of small-to-moderate effects (Cohen's d = 0.49), appropriate for an exploratory study.

### A. Descriptive Statistics

Per-joke counts and proportions were computed across all participants; funniness ratings varied by humor type. Aggressive jokes received the highest proportion of "funny = yes" responses ($M \approx 0.61$, $SD \approx 0.19$), followed by Affiliative ($M \approx 0.48$, $SD \approx 0.21$), Self-Enhancing ($M \approx 0.52$, $SD \approx 0.20$), and Self-Defeating ($M \approx 0.39$, $SD \approx 0.18$), see Fig. 1A. Appropriateness ratings were generally high across humor types ($M \approx 0.81$, $SD \approx 0.17$), though political jokes displayed lower appropriateness ($M \approx 0.68$, $SD \approx 0.22$) compared with person-themed jokes. Mean funniness also differed across semantic categories: person jokes ($M \approx 0.49$, $SD \approx 0.18$) were rated slightly funnier than political jokes ($M \approx 0.44$, $SD \approx 0.19$).

### B. Inferential Results

To analyze the effect of Content (Person vs. Politics) and Humor Type (Affiliative, Self-Enhancing, Aggressive, Self-Defeating) on the binary outcome variables Funny and Appropriate, a series of Generalized Linear Mixed Models (GLMM) with a binomial distribution and a log-link function were used. Student ID was included as a random intercept to account for the non-independence of evaluations made by the same student. A separate analysis was performed for language preference.

### C. Impact on Funniness and Appropriateness

The results are reported using the Wald $\chi^2$ statistic for the fixed effects.

**Funniness (Yes/No) GLMM:** A GLMM was conducted to examine the main effects of Content and Humor Type, as well as their interaction, on the likelihood of a joke being rated as funny. There was no statistically significant main effect of Content (Wald $\chi^2(1) = 0.54$, $p = 0.46$), indicating that people jokes and political jokes were rated as funny at comparable rates. In contrast, Humor Type showed a statistically significant main effect (Wald $\chi^2(3) = 14.12$, $p = 0.003$), suggesting that some types of humor were more likely to be rated as funny than others. The interaction between Content and Humor Type was not statistically significant (Wald $\chi^2(3) = 2.37$, $p = 0.50$). Post-hoc comparisons revealed that Affiliative and Aggressive jokes were rated as significantly funnier than Self-Enhancing and Self-Defeating jokes ($p < 0.05$), with Aggressive jokes showing the highest odds of being rated funny (OR = 1.68, 95% CI = [1.21, 2.33]) relative to Self-Defeating jokes, see Fig. 1B.

**Appropriateness (Yes/No) GLMM:** A GLMM was conducted to examine the main effects of Content and Humor Type and their interaction on the likelihood of a joke being rated as appropriate. There was a statistically significant main effect of Content (Wald $\chi^2(1) = 9.87$, $p = 0.002$), with Person jokes being more likely to be rated as appropriate than Political jokes (OR = 1.57, 95% CI = [1.17, 2.10]). Humor Type also showed a significant main effect (Wald $\chi^2(3) = 8.16$, $p = 0.043$), whereas the interaction between Content and Humor Type was not significant (Wald $\chi^2(3) = 1.39$, $p = 0.707$). Post-hoc analyses indicated that Aggressive and Affiliative jokes were rated as significantly more appropriate than Self-Enhancing and Self-Defeating jokes, though the effect was less pronounced than for funniness, see Fig. 1C.

### D. Bayesian Logistic Regression Results

To evaluate the effects of Content and Humor Type on participants' judgements of funniness and appropriateness, we conducted Bayesian logistic regressions. Evidence for the null hypothesis (H0) versus the alternative hypothesis (H1) was quantified using Bayes Factors (BF10), where BF10 > 1 indicates evidence for H1 (the presence of an effect), and BF10 < 1 indicates evidence for H0 (absence of an effect).

For *Funniness* ratings, the model provided strong evidence for the absence of a Content effect (BF10 = 0.09), but strong evidence for an effect of Humor Type (BF10 = 8.16). For *Appropriateness* ratings, the model indicated moderate evidence for an effect of Content (BF10 = 6.43), and anecdotal evidence for a small effect of Humor Type (BF10 = 2.11), following Jeffreys' interpretive scale [36].

### E. Impact on Language Preference

To analyze language preference for each joke, a Multinomial Logistic Regression was conducted, with Content, Humor Type, and the student's self-reported Fluency and Joking Preference languages as independent variables, and the preferred language (FR/EN/Both) as the dependent variable.

There was a statistically significant main effect of Content (Wald $\chi^2(2) = 10.32$, $p = 0.006$) on language preference. Specifically, Politics jokes were more likely to be preferred in FR or Both, whereas Person jokes were heavily favored in EN. In contrast, Humor Type did not show a significant main effect on language preference (Wald $\chi^2(6) = 6.45$, $p = 0.375$), and the interaction between Content and Humor Type was also not statistically significant (Wald $\chi^2(6) = 3.98$, $p = 0.679$), see Fig. 1D.

The student's self-reported Fluency language significantly predicted their preferred language for the joke (Wald $\chi^2(4) = 14.89$, $p = 0.005$), with students fluent in EN more likely to prefer EN, and similarly for FR or Both. Additionally, the student's self-reported Joking Preference language was a significant predictor of preferred language (Wald $\chi^2(4) = 13.06$, $p = 0.011$), such that students who preferred to make jokes in a particular language were more likely to select that language when evaluating the jokes.

## V. Discussion

Humor is crucial in human group interpersonal relationships, and computational humor allows us to apply it in group settings for human-robot interactions. As stated by [37], group-level processes are fundamentally different from individual-level processes, and measuring them in a strict scientific environment presents challenges that one-robot-to-one-human measurements do not. In this study, we examined how joke content, humor type, and individual language preferences influence evaluations of funniness, appropriateness, and language choice on a joke being told by a robot in a group setting. Regarding funniness, humor type emerged as a significant predictor, whereas content did not. Post-hoc analyses indicated that aggressive and affiliative jokes were rated as significantly funnier than self-enhancing and self-defeating jokes. For appropriateness, the GLMM revealed a significant main effect of content ($p = 0.002$), with personal jokes being rated as significantly more appropriate than political jokes. Humor Type also showed a significant effect ($p = 0.043$), with aggressive and affiliative humor rated slightly more appropriate. Additionally, with political jokes more likely to be preferred in French or both languages, taking into account that they were targeting the political situation in the francophone country, while personal jokes were strongly preferred in English.

The results partially support our hypotheses and highlight the complex interplay between humor style, topic, and social context in robotic humor. Consistent with H1, humor type significantly influenced perceived funniness, with affiliative and aggressive jokes receiving the highest funniness ratings. This finding aligns with prior research suggesting that affiliative humor promotes group bonding, while aggressive humor can be perceived as playful teasing within established social groups [3], [30]. Supporting H2, humor type influences perceived appropriateness, and supporting H3, appropriateness judgments were primarily shaped by joke content; person-related jokes were perceived as more appropriate than political jokes. Political humor often involves criticism of authority or sensitive societal topics, which may lead listeners to evaluate such jokes more cautiously when delivered by a robot [7].

These findings suggest that when robots engage in humor within group contexts, the topic of the joke may be more important for maintaining social acceptability than the humor style itself. Finally, language preferences varied depending on both participants' linguistic background and the joke content, providing partial support for H5. This result highlights the importance of cultural and linguistic context in humor perception and suggests that multilingual robotic systems may benefit from adapting humor delivery to users' language preferences.

Overall, these results suggest that the style of humor primarily drives perceptions of funniness, while both content and personal language factors influence appropriateness and language preference. The absence of significant interactions in both GLMM analyses indicates that Content and Humor Type largely operate independently in shaping joke evaluations.

### *A. Implications*

The current findings have several implications for the design of computational humor and HRI in group settings. First, the observed effects of humor type and content on funniness and appropriateness provide empirical guidelines for developing algorithms capable of generating jokes that are engaging while remaining socially acceptable. These insights can inform computational humor systems to balance humor impact with social sensitivity, particularly when robots interact with human groups. In collaborative and social contexts, robots could leverage humor to facilitate group cohesion, mediate conflicts, or introduce new members. For instance, robots operating in collaborative environments such as classrooms or workplaces may benefit from prioritizing affiliative humor while carefully moderating political or socially sensitive topics. The choice of personal versus political jokes highlights the importance of stress-testing the social limits of humor to ensure that robots are treated as acceptable social agents rather than disruptive outsiders. Furthermore, our findings underscore the importance of considering language and culture in both humor generation and evaluation. Multilingualism appears to shape openness to humorous content and social engagement, suggesting that algorithms and robots should adapt humor dynamically based on user characteristics and contextual factors.

### *B. Limitations*

We acknowledge some limitations in the study design, such as the limited number of jokes per category. Each humor type and each topic category was represented by a small number of AI-generated jokes. This limits the ecological validity of the stimuli and constrains the generalizability of the humor effects. Humor is highly sensitive to writing style, timing, and cultural references, and a small set of examples may not fully capture the natural variation within each humor style. Moreover, some jokes inevitably encode stylistic or semantic features that converge with humor type or topic. Although our mixed-effects modeling approach partially accounts for item-level variability, the possibility of residual confounds remains because jokes were not systematically counterbalanced across topics and humor types. In addition, future studies could benefit from a gender analysis as well.

Another limitation constitutes the potential ceiling effects on appropriateness ratings since it was rated at or near the ceiling for many jokes, reducing sensitivity to detect differences and inflating uncertainty in the posterior estimates. This may explain why some humor types (e.g., aggressive) appear unexpectedly appropriate in the Bayesian model. A more sensitive or multi-dimensional appropriateness measure (e.g., social acceptability, offensiveness, workplace-fit) would provide a more nuanced understanding.

A further limitation of this study concerns the coupling of joke generation and embodiment. In our experimental setup, AI-generated jokes were delivered by a robot, which means that the effects of content generation (AI vs. human) and embodiment (robot vs. human) cannot be fully disentangled. To more precisely isolate these factors, future work could employ a factorial design in which robots deliver both AI- and human-generated jokes, while humans deliver jokes produced by either humans or AI. However, the goal of the present exploratory study was to investigate a scenario in which humor is generated and performed by a fully artificial system. By leveraging LLMs for joke generation and deploying them through an embodied robotic platform, we examined how participants respond to humor produced and delivered by an autonomous artificial agent.

## VI. Conclusion

This study provides initial evidence that humor style, topic, and linguistic context jointly shape how robot-generated jokes are perceived in group settings. These findings contribute to the design of socially aware humor systems capable of adapting to cultural and contextual factors in human-robot interaction.

## Acknowledgements

Special thanks to Region Bretagne and Finistere, all IEVA MSc students participating in the study, and Julien Guyader for technical support.

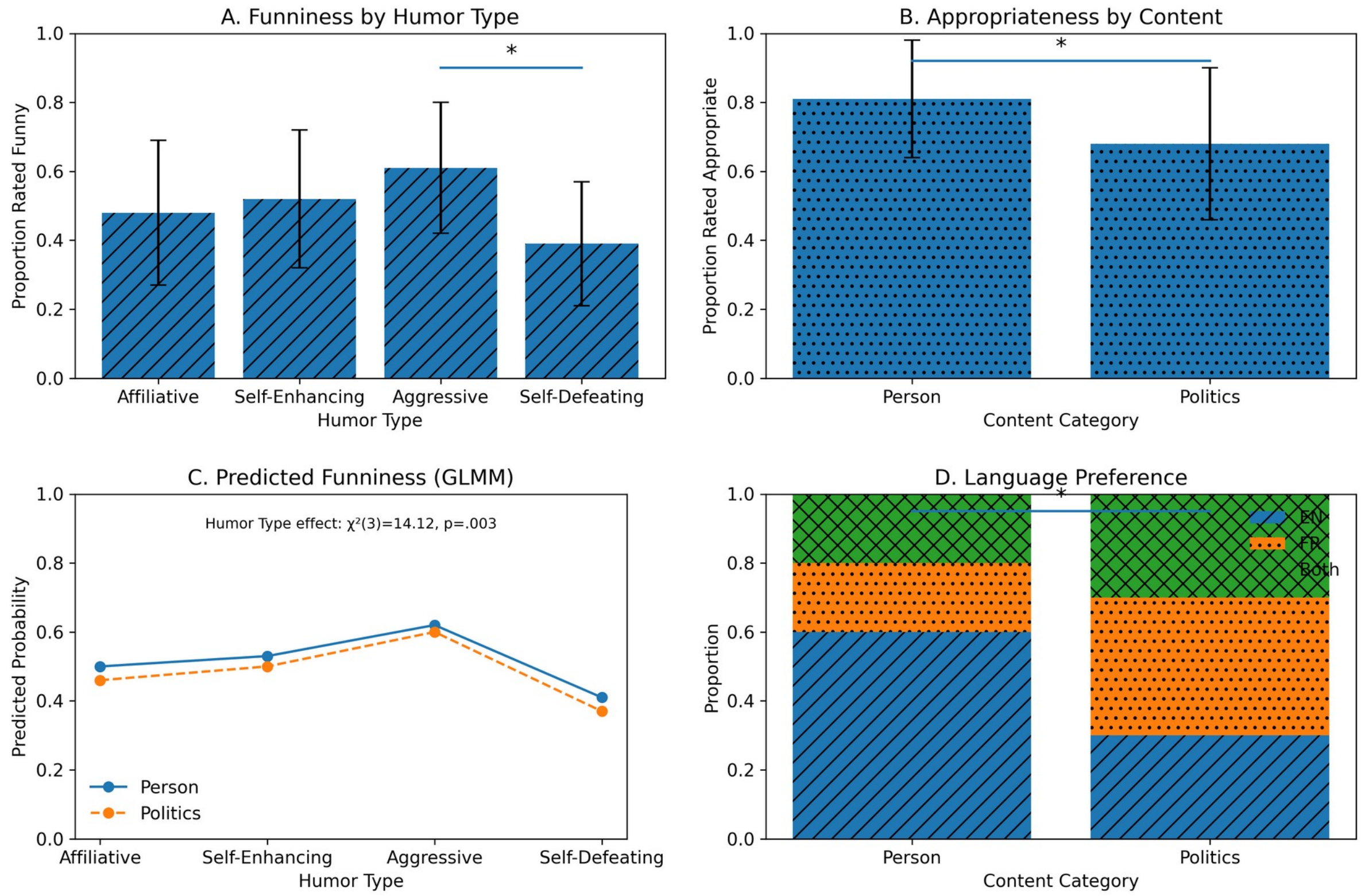

Fig. 1: Overview of descriptive and inferential results. (A) Mean proportion of jokes rated funny across humor types. Aggressive jokes were rated significantly funnier than self-defeating jokes. (B) Mean proportion rated appropriate by content category, showing higher appropriateness ratings for person-themed jokes. (C) Predicted probabilities of funniness from the GLMM, illustrating the significant main effect of humor type. (D) Distribution of preferred language for jokes by content category. Asterisks indicate statistically significant effects ($p < .05$). Error bars represent ±1 SD.

| Humor Type | Element | Description / Example |
|---|---|---|
| Affiliative | Definition | Enhance relationships and make others laugh in a socially friendly way. |
| | Example (EN) | " Nom says her humanoid robot is learning Greek culture — it already refuses to follow instructions unless you ask twice and offer a coffee first." |
| | Example (FR) | "Nom dit que son robot humanoïde apprend la culture grecque : il ne fait rien tant qu'on ne lui a pas demandé deux fois… et offert un café." |
| Self-Enhancing | Definition | A coping mechanism used to maintain a positive outlook. |
| | Example (EN) | "When her robot crashed right before a demo, Nom just shrugged and said, 'At least one of us is allowed a meltdown without a PhD defense." |
| | Example (FR) | "Quand son robot s'est écrasé juste avant une démonstration, Nom a haussé les épaules et a dit : Au moins, l'une de nous a le droit à un plantage sans avoir à fournir de thèse de doctorat." |
| Aggressive | Definition | Humor used to mock, tease, or put others down. |
| | Example (EN) | "Nom told her robot it could never replace her. Not because it lacks empathy — but because even a neural net couldn't survive Brest's weather and university administration." |
| | Example (FR) | "Nom a dit à son robot qu'il ne pourrait jamais la remplacer. Pas parce qu'il manque d'empathie… mais parce qu'aucune IA ne survivrait au climat de Brest et à l'administration universitaire." |
| Self-Defeating | Definition | Humor that undermines oneself in order to gain social approval. |
| | Example (EN) | "Nom says her robot is smarter than she is — it learned to play volleyball faster, remembers to charge itself, and never dyed its hair pink by accident." |
| | Example (FR) | "Nom dit que son robot est plus intelligent qu'elle : il a appris le volley plus vite, il se recharge tout seul, et il n'a jamais eu les cheveux roses par erreur." |

TABLE I: Definitions and examples of humor types in English and French within the personal context. Nom corresponds to name from the person of the group

| **Humor Type** | **Example** |
|---|---|
| Affiliative | **EN:** "They keep announcing a new Prime Minister in France like it's a pop-song release. I myself have had more stable relationships—with my houseplant. Meanwhile the French public looks at the latest PM and says: 'Nice to meet you… oh wait, you're already resigning?' At least my succulent doesn't ask for its own resignation letter."<br>**FR:** "Ils annoncent un nouveau Premier ministre en France comme s'il s'agissait de la sortie d'un tube pop. Moi-même, j'ai eu des relations plus stables — avec ma plante verte. Pendant ce temps, les Français regardent le dernier Premier ministre et disent : 'Enchanté… ah non, vous démissionnez déjà ?' Au moins, ma succulente ne demande pas sa propre lettre de démission." |
| Self-Enhancing | **EN:** "Following French politics lately has actually been helping my personal growth. Every time a new Prime Minister appears and disappears in record time, I remind myself: 'See? At least my projects last longer than that.' It's surprisingly motivating—if they can survive a week in office, I can definitely survive my deadlines."<br>**FR:** "Suivre la politique française ces derniers temps m'aide presque à me développer personnellement. Chaque fois qu'un nouveau Premier ministre apparaît puis disparaît en un temps record, je me dis : 'Tu vois ? Au moins mes projets durent plus longtemps que ça.' C'est étonnamment motivant — s'ils peuvent survivre une semaine au pouvoir, je peux sûrement survivre à mes deadlines." |
| Aggressive | **EN:** "Have you followed what's happening in French politics lately? It's like watching someone try to rob the Louvre Museum—and then the government says 'Oops, wrong door,' and still comes out with the Mona Lisa."<br>**FR:** "Vous avez suivi ce qui se passe dans la politique française ces derniers temps ? C'est comme regarder quelqu'un essayer de cambrioler le Louvre — puis le gouvernement dire : 'Oups, mauvaise porte,' et quand même repartir avec la Joconde." |
| Self-Defeating | **EN:** "I'm no expert in French politics—but at this point I feel like I could be Prime Minister too. I mean, I'm available, I've got opinions, and statistically one of us might outlast the next three in office. If anything, I think I'd fit in just by showing up and staying quiet for a bit longer than the guy before me."<br>**FR:** "Je ne suis pas un expert de la politique française — mais à ce stade, j'ai l'impression que je pourrais être Premier ministre moi aussi. Je veux dire, je suis disponible, j'ai des opinions, et statistiquement l'un de nous pourrait bien survivre aux trois prochains au pouvoir. Au fond, je pense que je m'intégrerais juste en arrivant et en restant silencieux un peu plus longtemps que le type avant moi." |

TABLE II: Examples of the four humor styles used in the study with English (EN) and French (FR) versions.